%% file: main.tex
\documentclass[11pt]{article}

\usepackage[final]{acl}
\usepackage{times}
\usepackage{latexsym}
\usepackage[T1]{fontenc}
\usepackage[utf8]{inputenc}
\usepackage{microtype}
\usepackage{hyperref}
\usepackage{url}
\usepackage{booktabs}
\usepackage{xspace}
\usepackage{graphicx}
\usepackage{textcomp}
\usepackage{multicol}
\usepackage{multirow}
\usepackage{enumitem}
\usepackage{pifont}
\usepackage{amsmath,amssymb}
\usepackage{subfigure}
\usepackage{float}
\usepackage{tikz}
\usepackage{subcaption}
\usepackage{lineno}
\usepackage{makecell}
\usepackage{CJKutf8}
\usepackage{inconsolata}
\usepackage{graphicx}
\usepackage[table]{xcolor}
\usepackage[normalem]{ulem}
\usepackage{bbm}
\usepackage{graphicx}
\usepackage{tabularx}  % X column
\usepackage{makecell}  % \makecell{...\\...}
\usepackage{array}     % \arraybackslash (안 잡히면 꼭 필요)
\usepackage{amssymb}

\definecolor{darkblue}{rgb}{0, 0, 0.5}
\hypersetup{colorlinks=true, citecolor=darkblue, linkcolor=darkblue, urlcolor=darkblue}

\definecolor{mycolor}{RGB}{33, 95, 154}
\definecolor{custom_red}{RGB}{228, 54, 54}

\title{Gap-K\%: Measuring Top-1 Prediction Gap for Detecting Pretraining Data}

\author{
  Minseo Kwak \quad
  Jaehyung Kim \\
  Yonsei University \\
  \texttt{\{rhkralstj103,jaehyungk\}@yonsei.ac.kr}
}

\begin{document}
\maketitle

\input{0_abstract}
\input{1_intro}
\input{2_related}
\input{3_method}

\input{4_experiments}
\input{5_conclusion}

\bibliography{custom}
\appendix
\input{6_appendix}

\end{document}

%% file: 0_abstract.tex
\begin{abstract}

The opacity of massive pretraining corpora in Large Language Models (LLMs) raises significant privacy and copyright concerns, making pretraining data detection a critical challenge.
Existing state-of-the-art methods typically rely on token likelihoods, yet they often overlook the gap between the target token and the model's top-1 prediction, as well as local correlations between adjacent tokens.
In this work, we propose Gap-K\%, a novel pretraining data detection method grounded in the optimization dynamics of LLM pretraining. 
By analyzing the next-token prediction objective, we observe that discrepancies between the model’s top-1 prediction and the target token induce strong gradient signals, which are explicitly penalized during training.
Motivated by this, Gap-K\% leverages the log probability gap between the top-1 predicted token and the target token, incorporating a sliding window strategy to capture local correlations and mitigate token-level fluctuations. 
Extensive experiments on the WikiMIA and MIMIR benchmarks demonstrate that Gap-K\% achieves state-of-the-art performance, consistently outperforming prior baselines across various model sizes and input lengths\footnote{Code:  \url{https://github.com/meaoww/gap-k}.}.

\end{abstract}

%% file: 1_intro.tex
\section{Introduction}

%1.
Large Language Models (LLMs) have recently demonstrated remarkable capabilities, including understanding, reasoning, and generation tasks \citep{openai2024gpt4o, comanici2025gemini}. 
A key to this success is pretraining on massive-scale data, mostly collected through web crawling. 
However, the details of these pretraining corpora remain largely undisclosed for many state-of-the-art models \citep{yang2025qwen3, grattafiori2024llama}. 
This lack of transparency raises significant concerns, as pretraining data may contain Personally Identifiable Information (PII) \citep{lukas2023analyzing} or copyrighted content \citep{rahman2023beyond}, leading to ethical and legal issues. 
Moreover, if benchmark datasets are unknowingly included in the pretraining data, model performance would be inflated, leading to unfair comparisons \citep{oren2023proving}.

%2.
For these reasons, \textit{pretraining data detection} \citep{shi2024detecting} has emerged as a critical problem; this problem is an instance of membership inference attack (MIA) \citep{shokri2017membership}, which aims to determine whether a data point was in the training dataset. 
Existing state-of-the-art methods typically address this problem by exploiting token-level likelihoods. 
For instance, Min-K\% \citep{shi2024detecting} assumes that non-training samples contain outlier tokens with low log probabilities, thus utilizing the average log probability of the bottom K\% tokens as a detection score.
Min-K\%++ \citep{zhang2025mink} further improves this method by normalizing token log probabilities relative to the mean and standard deviation of the next-token distribution.
However, these approaches treat each token independently, failing to exploit the local correlations between adjacent tokens.
Moreover, existing methods overlook the gap between the target token and the model's top-1 prediction, thereby missing informative signals induced by the training process.

%3.
In this work, we propose \textbf{Gap-K\%}, a novel pretraining data detection method grounded in the optimization dynamics of LLM pretraining. 
We leverage the insight that the next-token prediction objective explicitly forces the model to align its top-1 prediction with the ground truth.
Consequently, while training samples exhibit minimal difference from the top-1 prediction, unseen data frequently triggers \textit{confident mispredictions}, cases where the model strongly favors a plausible candidate other than the target.
Building on this observation, Gap-K\% quantifies the normalized gap between the log probabilities of the top-1 predicted token and the target token, effectively capturing such confident mispredictions. 
Furthermore, to exploit the local correlations within the token sequence, we apply a sliding window to the scores over adjacent tokens, thereby mitigating token-level fluctuations.

% 4.
We validate the effectiveness of Gap-K\% through experiments on two representative benchmarks, WikiMIA \citep{shi2024detecting} and MIMIR \citep{duan2024membership}. 
On WikiMIA, Gap-K\% consistently outperforms prior state-of-the-art methods across five evaluated models. 
In the original setting, averaged across models, Gap-K\% achieves absolute AUROC improvements of 9.7\% over the average of the existing baselines and 2.4\% over the strongest baseline, Min-K\%++. 
In the paraphrased setting, it achieves absolute AUROC gains of 5.7\% over the baseline average and 1.7\% over Min-K\%++.
On the more challenging MIMIR benchmark, Gap-K\% attains the highest average performance across Pythia models ranging from 1.4B to 12B parameters.

In summary, our contributions are as follows:

\begin{itemize}[leftmargin=3.5mm, itemsep=3pt, topsep=3pt, parsep=0.5pt]  
\item[$\circ$] {We identify that the gaps between top-1 predictions and target tokens as an effective detection signal, grounded in the optimization dynamics of next-token prediction.}
\item[$\circ$] {We propose Gap-K\%, a novel pretraining data detection method that leverages top-1 prediction gaps and aggregates token-level signals to capture local correlations between adjacent tokens.}
\item[$\circ$]{We achieve state-of-the-art performance on WikiMIA and MIMIR benchmarks across various model sizes and input lengths.}
\end{itemize}

%% file: 2_related.tex
\section{Related Works} 

\paragraph{Membership inference attacks.} 
Membership Inference Attacks (MIAs) aim to determine whether a given data sample was included in the training set. Samples included in the training data are referred to as members, while samples not included are non-members.
In the context of LLMs, MIAs have been widely used for quantifying memorization \citep{carlini2022quantifying} and privacy risks \citep{mireshghallah2022quantifying,steinke2023privacy}, as well as for detecting data contamination \citep{oren2023proving} and exposure of copyrighted content \citep{duarte2024cop, meeus2024did}. 
MIAs can be categorized into reference-based and reference-free methods. Reference-based methods \citep{carlini2021extracting, mireshghallah2022quantifying, ye2022enhanced} use additional reference models that are trained on data from a similar distribution. However, obtaining reference models is costly and often impractical, especially for pretrained LLMs. Therefore, reference-free methods have gained attention recently. Reference-free methods typically rely on loss \citep{yeom2018privacy}, and are further improved by comparing it with losses on perturbed samples \citep{mattern2023membership} or calibrating using compression-based entropy \citep{carlini2021extracting}.

\paragraph{Pretraining data detection.} 
Pretraining data detection is a specific instance of MIAs, where the objective is to infer whether a given text is included in the pretraining corpus of an LLM. Compared to standard MIAs, detecting pretraining data is more challenging because the data is seen only a few times within a massive corpus, resulting in weak memorization signals. Moreover, the lack of access to the pretraining data distribution makes it difficult to employ reference models \citep{shi2024detecting}. Prior work \citep{shi2024detecting, zhang2025mink} has focused on token-level probabilities and treats tokens independently. In contrast, we consider sequential dependencies in text and leverage top-1 predictions, which have received little attention in prior work.

%% file: 3_method.tex
\section{Method}
\subsection{Preliminary}
\paragraph{Problem definition.} 
Let $\mathcal{M}$ be an autoregressive language model trained on an unknown dataset $\mathcal{D}$. 
Given a text sequence $\mathbf{x}= [x_1,\dots,x_N]$, the goal is to determine whether $\mathbf{x}$ is a member of the training set ($\mathbf{x} \in \mathcal{D}$) or not ($\mathbf{x} \notin \mathcal{D}$). 
Specifically, a detection method computes a membership score $s(\mathbf{x};\mathcal{M})$, predicting $\mathbf{x} \in \mathcal{D}$ if $s(\mathbf{x}; \mathcal{M})$ exceeds a threshold $\lambda$.
We adopt a gray-box setting, where the model's output logits and token probabilities can be accessed, but the model parameters and gradients are not available.

\paragraph{Detection with token probabilities.} 
Existing state-of-the-art methods, such as Min-K\% \citep{shi2024detecting} and Min-K\%++ \citep{zhang2025mink}, rely on the likelihood of tokens. 
Min-K\% computes a membership score by averaging log probabilities of the lowest $k\%$ tokens:
\begin{equation}
\text{Min-K}(\mathbf{x})
= \frac{1}{|\mathcal{I}_k(\mathbf{x})|}
\sum_{t \in \mathcal{I}_k(\mathbf{x})}
\log p(x_t \mid x_{<t}),
\label{eq:mink}
\end{equation}
where $\mathcal{I}_k(\mathbf{x})$
denotes the set of indices of the lowest $k\%$ probability tokens.
Min-K\%++ builds on the intuition that training tokens are located near local maxima of the likelihood landscape, and thus have higher probabilities relative to other candidates in the vocabulary. 
Min-K\%++ formulates this by normalizing log-likelihoods:
\begin{align}
    \text{Min-K++}(\mathbf{x}) = \frac{1}{|\mathcal{I'}_k(\mathbf{x})|} \sum_{t \in \mathcal{I'}_k(\mathbf{x})} z_{t}, \label{eq:minkpp}\\
    z_{t} = \frac{\log p(x_t \mid x_{<t}) - \mu_{t}}{\sigma_{t}},
\label{eq:minkpp_token}
\end{align}
where $\mu_{t} = \mathrm{E}_{z \sim p(\cdot \mid x_{<t})}[\log p(z \mid x_{<t})]$
is the mean of the next-token log probability and
$\sigma_{t} =
\sqrt{\mathrm{E}_{z \sim p(\cdot \mid x_{<t})}
[(\log p(z \mid x_{<t}) - \mu_{t})^2]}
$
is the standard deviation.
Here, $\mathcal{I}_k'(\mathbf{x})$ denotes the set of indices corresponding to the lowest $k\%$ values of $\{z_t\}$.

\subsection{Motivation and Insight}

To distinguish training data from non-training data effectively, we focus on the fundamental behavior of the next-token prediction objective; we hypothesize that \textit{the primary fingerprint of training is in the alignment between the model's top prediction and the ground truth.}

\paragraph{Gradient-level analysis.} Consider the cross-entropy loss at step $t$, $\ell_t = - \log p(y_t \mid x_{<t})$, where $y_t$ denotes the next token in the pretraining sequence. 
The gradient of the loss with respect to logit $s_t(v)$ for any token $v \in V$ is:
\begin{equation}
    \frac{\partial \ell_t}{\partial s_t(v)} = p(v \mid x_{<t}) - \mathbbm{1}[v = y_t].
\end{equation}
Crucially, the magnitude of the gradient for non-target tokens ($v \neq y_t$) is directly proportional to their probability $p(v \mid x_{<t})$. 
The token with the highest probability, $v_t^{\max} = \arg\max_{v} p(v \mid x_{<t})$, exerts the strongest gradient signal if it does not match the target $y_t$. 
Therefore, during the training process, the optimization algorithm aggressively penalizes cases where $v_t^{\max} \neq y_t$.

As a consequence of this optimization, for samples within the training set $\mathcal{D}$, the model learns to align its top-1 prediction with the target token by minimizing the gap between $\log p(v_t^{\max})$ and $\log p(y_t)$.
In contrast, for unseen data $\mathbf{x} \notin \mathcal{D}$, the model cannot rely on memorization and instead predicts the next token based on learned syntactic and semantic correlations.
As a result, the target token ($y_t$) may not coincide with the top-1 prediction ($v_t^{\max}$), leading to a gap between their log probabilities. Since such mismatches are explicitly penalized during training, these gaps tend to be smaller for training data and larger for unseen data.
Thus, we hypothesize that this top-1 prediction gap serves as an informative signal for inferring membership.

\subsection{Proposed Method: Gap-K\%}
\input{Figures/Figure1}
Building on the insight that training data exhibits minimal top-1 prediction gaps, we propose \textbf{Gap-K\%}, a method that quantifies this gap while accounting for sequential dependencies.

\input{Figures/Figure2}

\paragraph{Top-1 gap scoring.} 
First, we measure the token-level gap. For each token $x_t$, we compute the difference between its log probability and the maximum log probability over the vocabulary (\textit{i.e.}, the top-1 prediction). 
To account for the varying sharpness of the output distribution (\textit{e.g.}, flat vs. peaked distributions), we normalize this difference by the standard deviation $\sigma_t$ of the log probabilities, similar to Min-K\%++:
\begin{equation}
    g_t = \frac{\log p(x_t \mid x_{<t}) - \max\limits_{v \in V} \log p(v \mid x_{<t})}{\sigma_t}.\label{eq:gap_score}
\end{equation}
Here, $g_t$ is always \textit{non-positive}; a value close to 0 indicates that the target token $x_t$ has a log probability close to that of the top-1 prediction, while a large negative value indicates a large gap between the target token and the top-1 prediction (likely for non-training data).

\paragraph{Sequential smoothing.}
Membership signals in LLMs are rarely isolated to a single token; they often span phrases or sentences, exhibiting sequential consistency due to the memorization of continuous text segments. 
To capture this local correlation and mitigate token-level fluctuations, we apply a sliding window of size $w$ over the gap scores:
\begin{equation}
\bar{g}_t^{(w)}
=
\frac{1}{w}\sum_{i=0}^{w-1} g_{t+i}.
\end{equation}
As illustrated in Fig. \ref{fig:my_example}, this smoothing step effectively highlights regions where the model consistently aligns (or fails to align) with the input text.

\paragraph{Gap-K\% metric.}
Finally, as in Min-K\%, we focus on the \textit{worst-case segments} that provide the strongest counter-evidence for membership. 
We define the {Gap-K\%} score as the average of the lowest $k\%$ smoothed gap scores in the sequence:
\begin{equation}
\text{Gap-K}(\mathbf{x})
=
\frac{1}{|\tilde{\mathcal{I}}_{k}(\mathbf{x})|}
\sum_{t \in \tilde{\mathcal{I}}_{k}(\mathbf{x})} \bar{g}_t^{(w)},
\end{equation}
where $\tilde{\mathcal{I}}_{k}(\mathbf{x})$ denotes the set of token indices corresponding to the
bottom $k\%$ values of the smoothed scores $\{\bar{g}_t^{(w)}\}_{t=1}^{N-w+1}$.
A higher Gap-K\% score (closer to 0) implies that even the hardest-to-predict segments of the text exhibit a small top-1 prediction gap, indicating membership in the training data.
We provide empirical validation of this design choice in Appendix~\ref{app:token_selection}.

\input{Tables/Table1}

\subsection{Comparison with Min-K\%++}
To validate the advantage of Gap-K\% over Min-K\%++, we analyze the relationship between our scoring function $g_t$ and the Min-K\%++ score $z_t$. 
By expanding Eq.~\ref{eq:gap_score} using the definition of $z_t$ (Eq.~\ref{eq:minkpp_token}), we can express $g_t$ as:
\begin{equation}
g_t
=
z_t
-
\underbrace{\frac{\max\limits_{v \in V} \log p(v \mid x_{<t}) - \mu_t}{\sigma_t}}_{\Delta_t}.
\end{equation}
This formulation shows that our score differs from Min-K\%++ by an additional term $\Delta_t$, capturing the normalized gap between the top-1 log probability and the mean. 
It provides two critical insights. 

\paragraph{Penalizing confident errors.}
Both methods focus on outlier tokens (lowest $k\%$ scores), which typically occur when the target token $x_t$ differs from the top-1 prediction.
In such scenarios, Min-K\%++ ($z_t$) measures deviation solely from the mean, treating all low-probability tokens similarly regardless of the distribution shape.
In contrast, Gap-K\% ($g_t$) incorporates the confidence term $\Delta_t$.
Crucially, this allows our method to distinguish between \textit{uncertain predictions} where the distribution is flat, and \textit{confident mispredictions} where the model assigns high probability to an incorrect token, as illustrated in Fig. \ref{fig:fig2}.
By imposing a severe penalty on the latter, Gap-K\% leverages these confident mispredictions as strong counter-evidence against membership. Since training samples are optimized to minimize such gaps, detecting these gaps provides a more robust signal for identifying non-training data than relying on likelihood alone.

\paragraph{Mode vs. Mean.} 
From a statistical perspective, Min-K\%++ assumes that training tokens are located near local maxima of the likelihood landscape. 
However, it adopts an \textit{indirect} approach to capture this by measuring deviations from the \textit{mean} of the vocabulary distribution.
In contrast, Gap-K\% aligns the metric with this assumption by directly measuring the gap between the target token and the mode of the distribution.

%% file: Figures/Figure1.tex
\begin{figure}[t]
  \centering
  \includegraphics[width=\linewidth]{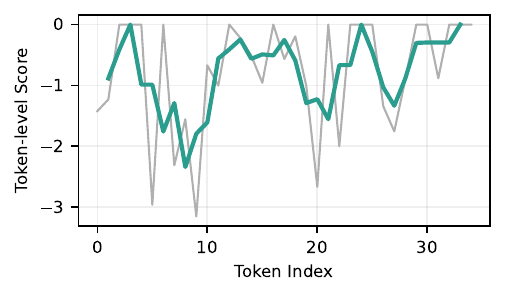}
  \vspace{-0.2in}
  \caption{Illustration of token-level scores before and after sequential smoothing. The gray curve shows raw token-level scores, while the green curve shows the smoothed scores.}
  \label{fig:my_example}
  \vspace{-0.15in}
\end{figure}

%% file: Figures/Figure2.tex
\begin{figure*}[t]
  \centering
  \includegraphics[width=0.75\textwidth]{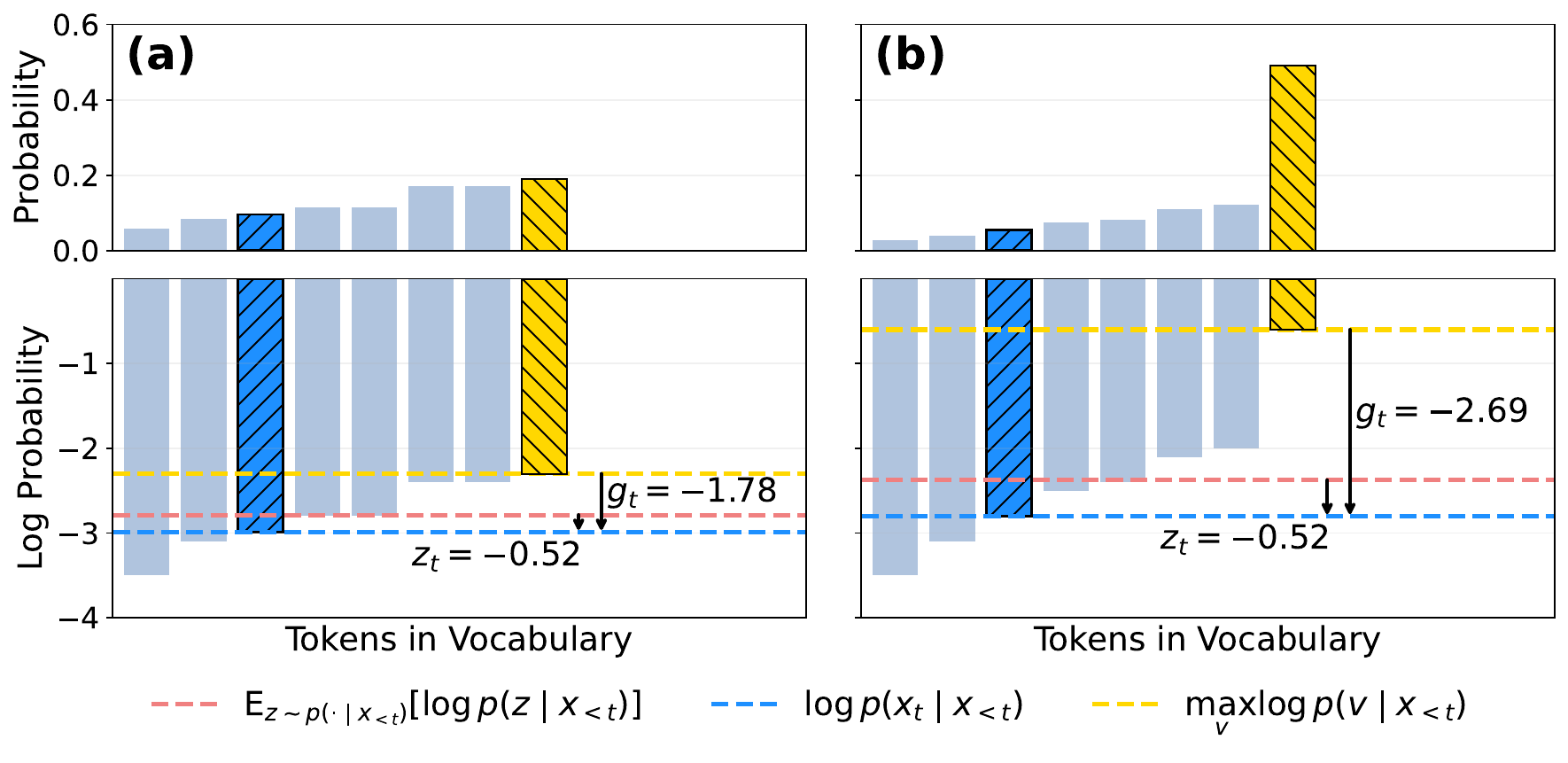}
  \vspace{-0.1in}
  \caption{Conceptual comparison between Min-K\%++ and Gap-K\% using a toy example with a small vocabulary size of 8. Inspired by the illustrative analysis in  \citet{zhang2025mink}, we compare the token-level scores of Min-K\%++ ($z_t$) and Gap-K\% ($g_t$) under two different next-token probability distributions. The $z_t$ and $g_t$ values annotated in the figure are normalized quantities. The blue hatched bar denotes the observed token $x_t$, while the yellow hatched bar indicates the top-1 token. In (a), the distribution is relatively flat, resulting in low-confidence incorrect predictions, whereas in (b) the model assigns high confidence to an incorrect top-1 token. While the Min-K\%++ score $z_t$ is identical in both cases, the Gap-K\% score $g_t$ distinguishes confident mispredictions from uncertain predictions by capturing the gap between the observed token and the top-1 prediction.}
  \label{fig:fig2}
\end{figure*}

%% file: Tables/Table1.tex
\begin{table*}[t]
\centering
\scriptsize
\setlength{\tabcolsep}{8pt}
\renewcommand{\arraystretch}{1.15}

\caption{AUROC results on WikiMIA. \textit{Ori.} and \textit{Para.} indicate the original and paraphrased settings, respectively. \textbf{Bold} numbers denote the best performance.}

\label{tab:main_wikimia}

\begin{tabular}{c l cc cc cc cc cc cc}
\toprule
& & \multicolumn{2}{c}{\textbf{Mamba-1.4B}}
  & \multicolumn{2}{c}{\textbf{Pythia-6.9B}}
  & \multicolumn{2}{c}{\textbf{Pythia-12B}}
  & \multicolumn{2}{c}{\textbf{LLaMA-13B}}
  & \multicolumn{2}{c}{\textbf{LLaMA-65B}}
  & \multicolumn{2}{c}{\textbf{Average}} \\
\cmidrule(lr){3-4}\cmidrule(lr){5-6}\cmidrule(lr){7-8}\cmidrule(lr){9-10}\cmidrule(lr){11-12}\cmidrule(lr){13-14}
\textbf{Len.} & \textbf{Method}
& \textit{Ori.} & \textit{Para.}
& \textit{Ori.} & \textit{Para.}
& \textit{Ori.} & \textit{Para.}
& \textit{Ori.} & \textit{Para.}
& \textit{Ori.} & \textit{Para.}
& \textit{Ori.} & \textit{Para.} \\
\midrule

% ===================== Len 32 =====================
\multirow{6}{*}{32}
& Loss
& 61.0 & 61.3
& 63.8 & 64.1
& 65.4 & 65.6
& 67.5 & 68.0
& 70.8 & 71.9
& 65.7 & 66.2 \\
& Zlib
& 61.9 & 62.3
& 64.4 & 64.2
& 65.8 & 65.9
& 67.8 & 68.3
& 71.2 & 72.1
& 66.2 & 66.6 \\
& Neighbor
& 64.1 & 63.6
& 65.8 & 65.5
& 66.6 & 66.8
& 65.8 & 65.0
& 69.6 & 68.7
& 66.4 & 65.9 \\
& Min-K\%
& 63.3 & 62.9
& 66.3 & 65.1
& 68.1 & 67.2
& 66.8 & 66.2
& 70.6 & 70.1
& 67.0 & 66.3 \\
& Min-K\%++
& 66.4 & 65.7
& 70.3 & 67.6
& 72.2 & 69.4
& 84.4 & 82.7
& 85.3 & 81.6
& 75.7 & 73.4 \\
\rowcolor{gray!15}
& Gap-K\%
& \textbf{69.2} & \textbf{67.2}
& \textbf{71.4} & \textbf{68.1}
& \textbf{73.7} & \textbf{70.2}
& \textbf{86.8} & \textbf{83.7}
& \textbf{88.0} & \textbf{82.5}
& \textbf{77.8} & \textbf{74.3} \\
\midrule

% ===================== Len 64 =====================
\multirow{6}{*}{64}
& Loss
& 58.2 & 56.4
& 60.7 & 59.3
& 61.9 & 60.0
& 63.6 & 63.1
& 67.9 & 67.9
& 62.5 & 61.3 \\
& Zlib
& 60.4 & 59.1
& 62.6 & 61.6
& 63.5 & 62.1
& 65.3 & 65.3
& 69.1 & 69.5
& 64.2 & 63.5 \\
& Neighbor
& 60.6 & 60.6
& 63.2 & 63.1
& 62.6 & 62.8
& 64.1 & 64.7
& 69.6 & 69.5
& 64.0 & 64.1 \\
& Min-K\%
& 61.7 & 58.0
& 65.0 & 61.1
& 66.5 & 62.5
& 66.0 & 63.5
& 69.9 & 66.8
& 65.8 & 62.4 \\
& Min-K\%++
& 67.2 & 62.2
& 71.6 & 64.2
& 72.6 & 65.1
& 84.3 & 78.8
& 83.5 & 74.3
& 75.8 & 68.9 \\
\rowcolor{gray!15}
& Gap-K\%
& \textbf{69.9} & \textbf{64.3}
& \textbf{73.3} & \textbf{66.7}
& \textbf{74.8} & \textbf{67.1}
& \textbf{87.2} & \textbf{81.2}
& \textbf{86.7} & \textbf{76.7}
& \textbf{78.4} & \textbf{71.2} \\
\midrule

% ===================== Len 128 =====================
\multirow{6}{*}{128}
& Loss
& 63.3 & 62.7
& 65.1 & 64.7
& 65.8 & 65.4
& 67.8 & 67.2
& 70.8 & 70.2
& 66.6 & 66.0 \\
& Zlib
& 65.6 & 65.3
& 67.6 & \textbf{67.4}
& 67.8 & 67.9
& 69.7 & 69.6
& 72.2 & \textbf{72.2}
& 68.6 & 68.5 \\
& Neighbor
& 64.8 & 62.6
& 67.5 & 64.3
& 67.1 & 64.3
& 68.3 & 64.0
& 73.7 & 70.3
& 68.3 & 65.1 \\
& Min-K\%
& 66.8 & 64.4
& 69.5 & 67.0
& 70.7 & 68.5
& 71.5 & 68.6
& 73.8 & 70.5
& 70.5 & 67.8 \\
& Min-K\%++
& 67.7 & 63.3
& 69.8 & 65.9
& 71.8 & 67.7
& 83.8 & 76.2
& 80.8 & 70.0
& 74.8 & 68.6 \\
\rowcolor{gray!15}
& Gap-K\%
& \textbf{71.2} & \textbf{67.4}
& \textbf{71.5} & 66.2
& \textbf{75.0} & \textbf{69.5}
& \textbf{85.7} & \textbf{79.4}
& \textbf{83.6} & 70.7
& \textbf{77.4} & \textbf{70.6} \\
\bottomrule
\end{tabular}
\end{table*}

%% file: 4_experiments.tex
\section{Experiments}
\input{Tables/Table2}
In this section, we conduct comprehensive experiments to evaluate the proposed Gap-K\%, designed to address the following research questions:
\begin{itemize}[leftmargin=3.5mm, itemsep=3pt, topsep=3pt, parsep=0.5pt]
    \item[$\circ$] \textbf{RQ1:} How effectively can Gap-K\% detect pretraining data compared to existing baselines? (Table \ref{tab:main_wikimia}, \ref{tab:main_mimir} and Fig.~\ref{fig:fig3})
    \item[$\circ$] \textbf{RQ2:} Is the top-1 prediction gap empirically supported as a detection signal? (Table~\ref{tab:gap_threshold})
    \item[$\circ$] \textbf{RQ3:} How do the components and hyperparameters of Gap-K\% affect detection performance? (Table~\ref{tab:sequential_locality},\ref{tab:ablation} and Fig.~\ref{fig:window_size},\ref{fig:ratio})
    \item[$\circ$] \textbf{RQ4:} Does Gap-K\% generalize to recent LLMs and remain robust under adversarial paraphrasing? (Table~\ref{tab:recent_models},\ref{tab:dipper})
    \item[$\circ$] \textbf{RQ5:} How does Gap-K\% work differently from Min-K\%++? (Fig.~\ref{tab:recent_models})

\end{itemize}

\subsection{Setup}
\paragraph{Benchmarks.}
We evaluate our method on WikiMIA \citep{shi2024detecting} and MIMIR \citep{duan2024membership}, two representative benchmarks for pretraining data detection. 
WikiMIA is constructed based on Wikipedia event pages and assigns training and non-training labels based on timestamps. 
WikiMIA provides both original and paraphrased versions. 
Since detection difficulty may vary with input length, WikiMIA includes length-based splits. In our experiments, we evaluate the length splits of 32, 64, and 128 for both the original and paraphrased settings, following \citet{zhang2025mink}.
MIMIR is a more challenging benchmark, designed to have minimal distributional differences between training and non-training samples. 
MIMIR is constructed from the Pile dataset \citep{gao2020pile} and consists of seven domains: English Wikipedia, GitHub, Pile-CC, PubMed Central, arXiv, DM Mathematics, and HackerNews.

\paragraph{Models.}
Since WikiMIA is constructed from Wikipedia, which is included in the pretraining data of many LLMs, we evaluate Mamba-1.4B \citep{gu2024mamba}, Pythia-6.9B and Pythia-12B \citep{biderman2023pythia}, LLaMA-13B and LLaMA-65B \citep{touvron2023llama}. For MIMIR, which is designed for models trained on the Pile dataset, we follow \citet{duan2024membership} and evaluate Pythia-160M, 1.4B, 2.8B, 6.9B, and 12B.

\paragraph{Metrics.}
We evaluate the performance using the Area Under the ROC curve (AUROC) as our primary evaluation metric.
AUROC captures the overall trade-off between the true positive rate (TPR) and false positive rate (FPR) across different thresholds. 
We also report the true positive rate at a low false positive rate (TPR@5\%FPR).

\paragraph{Baselines.}
We compare our method with five state-of-the-art baselines: (1) \textit{Loss} \citep{yeom2018privacy} directly uses the input loss. (2) \textit{Zlib} \citep{carlini2021extracting} calibrates the loss using Zlib compression entropy. 
(3) \textit{Neighbor} \citep{mattern2023membership} perturbs the input text using a pretrained masked language model and compares the loss of the original sample with the average loss of perturbed samples. 
(4) \textit{Min-K\%} \citep{shi2024detecting} computes the average of the lowest $k$\% token probabilities. 
(5) \textit{Min-K\%++} \citep{zhang2025mink} extends Min-K\% by normalizing token-level log probabilities.

\paragraph{Implementation details.}
We use \(k = 20\%\) for Min-K\% following the setting in the original paper \citep{shi2024detecting}. To ensure a fair comparison across methods, we fix \(k = 20\%\) for Min-K\%++ and Gap-K\%. For Gap-K\%, we set the window size to 6 for LLaMA-based models and 3 for other models, as these settings consistently showed stable and near-optimal performance across model sizes within each model family.

\input{Figures/Figure3}

\subsection{Main Results}
\paragraph{WikiMIA results.}
Table~\ref{tab:main_wikimia} reports the AUROC scores on the WikiMIA benchmark. Overall, Gap-K\% consistently outperforms existing methods across diverse settings. Averaged over five models, in the original setting, Gap-K\% improves upon Min-K\%++ by 2.1\%, 2.6\%, and 2.6\% for input lengths of 32, 64, and 128, respectively. For the paraphrased setting, Gap-K\% also outperforms Min-K\%++ by 0.9\%, 2.3\%, and 2.0\%. Moreover, the performance gains of Gap-K\% generalize across a wide range of model sizes and architectures.
We further analyze TPR@5\%FPR, with full results reported in Appendix~\ref{app:fpr}.
Since Min-K\% and Min-K\%++ achieve the strongest AUROC among the baselines, we focus on these two in Fig.~\ref{fig:fig3}, which summarizes the average across input lengths.
Gap-K\% substantially outperforms Min-K\%++ in the original setting, with gains of 7.1\%, 7.9\%, and 3.0\% for input lengths of 32, 64, and 128, respectively. In the paraphrased setting, Gap-K\% also achieves consistent gains of 2.0\%, 3.4\%, and 2.1\% for the same input lengths.

\paragraph{MIMIR results.}
Table~\ref{tab:main_mimir} reports the AUROC results on the MIMIR benchmark. 
MIMIR is a particularly challenging setting, as member and non-member samples are drawn from nearly identical distributions, resulting in performance close to random guessing (0.5) for most methods.
This difficulty is further amplified in the Pile-CC subset, which consists of noisy web-crawled text with high entropy, weakening memorization signals even for training data (see Appendix~\ref{app:pile-cc} for details).
Despite these challenges, Gap-K\% achieves the strongest average performance on the MIMIR benchmark across 1.4B, 2.8B, 6.9B, and 12B models. Notably, Gap-K\% achieves higher average performance than Min-K\%++ across the five evaluated model sizes. The TPR@5\%FPR results are in Appendix \ref{app:fpr}.

\input{Tables/Table3}
\subsection{Additional Analyses}\label{sec:4.3}
We conduct additional analyses of Gap-K\% on the WikiMIA-64 dataset using Pythia-12B.

\paragraph{Empirical evidence of gap signal.} 
To examine how frequently confident mispredictions occur in non-training data compared to training data, we compute the normalized token-level gap $g_t$ (Eq.~\ref{eq:gap_score}) and measure the fraction of tokens whose gap magnitude $|g_t|$ exceeds a threshold $\tau$, where $\tau$ corresponds to a deviation of $\tau \cdot \sigma_t$ from the top-1 prediction. 
As shown in Table~\ref{tab:gap_threshold}, non-training data consistently exhibits a higher proportion of large-gap tokens across a wide range of thresholds. 
For example, at $\tau = 3$, such deviations occur in 39.9\% of non-training tokens, compared to 35.5\% in training data.
These results suggest that confident mispredictions are reduced during training, supporting the use of large top-1 gaps as a signal for distinguishing training from non-training data.

\paragraph{Impact of window size.}
Fig.~\ref{fig:window_size} presents the AUROC scores for window sizes ranging from 1 to 10 on LLaMA-13B and Pythia-12B. 
LLaMA-13B and Pythia-12B achieve their best performance at window sizes of 6 and 3, respectively, and the performance degrades for larger window sizes.  
As the window size increases, smoothing aggregates gap scores over a broader context, which may dilute localized high-gap regions that are indicative of non-training data. 
We further investigate the difference in optimal window size across model families in Appendix~\ref{app:window_size}.

\input{Figures/Figure4}
\input{Tables/Table4}
\paragraph{Effect of sequential locality.}
To examine whether the effectiveness of sequential smoothing stems from local token dependencies,
we compare three variants: (1) no smoothing, (2) smoothing applied after randomly shuffling the token order, and (3) sequential smoothing applied to the original token order. 
As shown in Table~\ref{tab:sequential_locality}, sequential smoothing leads to substantial improvements in AUROC only when the original token order is preserved.
In contrast, shuffling the token order beforehand yields minimal improvement. 
This observation indicates that the membership signal exhibits sequential locality, being distributed across contiguous token segments rather than isolated at individual tokens.

\input{Tables/Table5}
\input{Figures/Figure5}
\input{Figures/Figure6}

\paragraph{Ablation of key components.}
We ablate the two key differences from Min-K\%++: Top-1 gap score in place of mean-based normalization and sequential smoothing.
Table~\ref{tab:ablation} shows that neither modification alone is sufficient to achieve a large performance gain. 
Replacing the mean-based normalization with the Top-1 gap leads to a marginal drop in performance. 
On the other hand, applying sequential smoothing to the original Min-K\%++ score leads to an improvement in AUROC, indicating that smoothing is generally effective. 
Importantly, the largest performance gain is achieved when sequential smoothing is applied to the Top-1 gap score.
While smoothing alone stabilizes the score, the Top-1 gap provides a more discriminative per-token signal that becomes effective only when combined with sequential smoothing.

\paragraph{Effect of $k\%$.}
Fig.~\ref{fig:ratio} shows the effect of varying $k$ from 5\% to 50\%.
We observe that the performance peaks around $k=15\%$, suggesting that focusing on tokens with the largest log probability gaps yields the strongest detection signals.
Additionally, Gap-K\% consistently outperforms Min-K\% and Min-K\%++ across the 5\%–50\% range, regardless of the choice of $k$.

\input{Tables/Table6}
\paragraph{Generalization to recent models.}
We further evaluate Gap-K\% on recent LLMs, including LLaMA 3.1-8B \citep{grattafiori2024llama} and Gemma2-9B \citep{team2024gemma}, as well as their instruction-tuned variants.
Since the training data of recent models temporally overlaps with the original WikiMIA benchmark, we adopt WikiMIA-25 \citep{yi2026membership} (length 64 split) for this evaluation.
As shown in Table~\ref{tab:recent_models}, Gap-K\% consistently outperforms all baseline methods across both base and instruction-tuned models. 
These results further demonstrate that the proposed signal is not tied to a specific architecture or training setup and remains effective under modern training regimes, including instruction tuning and reinforcement learning from human feedback (RLHF).

\input{Tables/Table7}
\paragraph{Robustness to adversarial paraphrases.}
We evaluate the robustness of Gap-K\% under strong adversarial paraphrasing.
To the best of our knowledge, there is no established method specifically designed to generate paraphrases for evading pretraining data detection.
We adopt DIPPER \citep{krishna2023paraphrasing}, a paraphrasing framework originally proposed for attacking AI-generated text detectors.
We employ the strongest setting reported in the DIPPER paper, with lexical diversity $L=60$ and order diversity $O=60$.
As shown in Table~\ref{tab:dipper}, Gap-K\% outperforms baseline methods even under these challenging conditions, demonstrating its robustness to strong adversarial paraphrasing.

\paragraph{Qualitative example.} 
We analyze the behavior of token-level scores produced by Min-K\%++ and Gap-K\% to clarify how low-percentage selection mechanisms differ between the two methods. 
While the main quantitative experiments in prior sections are conducted on WikiMIA-64, we use WikiMIA-32 here for improved clarity in visualization. 
Fig.~\ref{fig:token_analysis} visualizes the token-level scores for a representative sample.
We observe that Min-K\%++ selects multiple low-scoring tokens distributed across the sequence, including both pronounced and minor low-score points. 
These selected tokens appear at isolated positions rather than forming segments. 
In contrast, Gap-K\% applies sequential smoothing before selection, which reflects score trends over neighboring tokens. 
As a consequence, the highlighted regions correspond to spans where scores remain relatively low across a short range, rather than isolated tokens. 
This suggests that smoothing allows Gap-K\% to capture membership signals at the level of local patterns, whereas Min-K\%++ operates on individual token values.

%% file: Tables/Table2.tex
\begin{table*}[t]
\centering
\scriptsize
\setlength{\tabcolsep}{3.2pt}
\renewcommand{\arraystretch}{1.15}

\caption{AUROC results on MIMIR under the 13-gram 0.8 overlap setting \citep{duan2024membership}. 
The best and second-best scores are highlighted in \textbf{bold} and \uline{underlined}, respectively. For the Pythia-12B model, results for the Neighbor method are not reported due to computational constraints, consistent with \citet{zhang2025mink}.}
\label{tab:main_mimir}

\begin{tabular}{l ccccc ccccc ccccc ccccc}
\toprule
& \multicolumn{5}{c}{\textbf{Wikipedia}}
& \multicolumn{5}{c}{\textbf{Github}}
& \multicolumn{5}{c}{\textbf{Pile CC}}
& \multicolumn{5}{c}{\textbf{PubMed Central}} \\
\cmidrule(lr){2-6}\cmidrule(lr){7-11}\cmidrule(lr){12-16}\cmidrule(lr){17-21}
\textbf{Method}
& 160M & 1.4B & 2.8B & 6.9B & 12B
& 160M & 1.4B & 2.8B & 6.9B & 12B
& 160M & 1.4B & 2.8B & 6.9B & 12B
& 160M & 1.4B & 2.8B & 6.9B & 12B \\
\midrule
Loss
& 50.2 & 51.3 & 51.8 & 52.8 & 53.5
& \uline{65.7} & 69.8 & 71.3 & 73.0 & 71.0
& 49.6 & 50.0 & 50.1 & 50.7 & 51.1
& 49.9 & 49.8 & 49.9 & 50.6 & 51.3 \\
Zlib
& \textbf{51.1} & 52.0 & 52.4 & 53.5 & 54.3
& \textbf{67.5} & \textbf{71.0} & \textbf{72.3} & \textbf{73.9} & 72.2
& 49.6 & 50.1 & \uline{50.3} & 50.8 & 51.1
& 49.9 & 50.0 & 50.1 & 50.6 & 51.2 \\
Neighbor
& \uline{50.7}  & 51.7 & 52.2 & 53.2 & /
& 65.3 & 69.4 & 70.5 & 72.1 & /
& 49.6 & 50.0 & 50.1 & 50.8 & /
& 47.9 & 49.1 & 49.7 & 50.1 & / \\
Min-K\%
& 48.8 & 51.0 & 51.7 & 53.1 & 54.2
& \uline{65.7} & \uline{70.0} & \uline{71.4} & \uline{73.3} & 72.2
& \uline{50.1} & \uline{50.5} & \textbf{50.5} & \uline{51.2} & 51.5
& \uline{50.3} & 50.3 & 50.5 & 51.2 & 52.3 \\
Min-K\%++
& 49.2 & \uline{53.1} & \uline{53.8} & \uline{56.1} & \uline{57.9}
& 64.7 & 69.6 & 70.9 & 72.8 & \uline{73.8}
& 49.7 & 50.0 & 49.8 & \uline{51.2} & \uline{51.7}
& 50.2 & \uline{50.8} & \uline{51.5} & \uline{52.8} & \uline{54.0} \\
\rowcolor{gray!15}
Gap-K\%
& 48.9 & \textbf{53.5} & \textbf{54.1} & \textbf{56.7} & \textbf{58.6}
& 64.0 & 69.6 & 71.0 & 73.1 & \textbf{74.1}
& \textbf{50.5} & \textbf{50.6} & \uline{50.3} & \textbf{51.6} & \textbf{52.0}
& \textbf{50.9} & \textbf{51.3} & \textbf{51.7} & \textbf{53.2} & \textbf{54.3} \\
\midrule
& \multicolumn{5}{c}{\textbf{ArXiv}}
& \multicolumn{5}{c}{\textbf{DM Mathematics}}
& \multicolumn{5}{c}{\textbf{HackerNews}}
& \multicolumn{5}{c}{\textbf{Average}} \\
\cmidrule(lr){2-6}\cmidrule(lr){7-11}\cmidrule(lr){12-16}\cmidrule(lr){17-21}
\textbf{Method}
& 160M & 1.4B & 2.8B & 6.9B & 12B
& 160M & 1.4B & 2.8B & 6.9B & 12B
& 160M & 1.4B & 2.8B & 6.9B & 12B
& 160M & 1.4B & 2.8B & 6.9B & 12B \\
\midrule
Loss
& \textbf{51.0} & \textbf{51.5} & 51.9 & 52.9 & 53.4
& 48.8 & 48.5 & 48.4 & 48.5 & 48.5
& 49.4 & 50.5 & 51.3 & 52.1 & 52.8
& 52.1 & 53.1 & 53.5 & 54.4 & 54.5 \\
Zlib
& 50.1 & 50.9 & 51.3 & 52.2 & 52.7
& 48.1 & 48.2 & 48.0 & 48.1 & 48.1
& 49.7 & 50.3 & 50.8 & 51.2 & 51.7
& \textbf{52.3} & 53.2 & 53.6 & 54.3 & 54.5 \\
Neighbor
& \uline{50.7} & \uline{51.4} & 51.8 & 52.2 & /
& 49.0 & 47.0 & 46.8 & 46.6 & /
& \textbf{50.9} & \textbf{51.7} & 51.5 & 51.9 & /
& 52.0 & 52.9 & 53.2 & 53.8 & / \\
Min-K\%
& 50.4 & \uline{51.4} & 52.1 & 53.4 & \uline{54.3}
& 49.3 & 49.3 & 49.1 & 49.2 & 49.2
& 50.6 & 51.2 & \uline{52.4} & 53.5 & 54.5
& \uline{52.2} & 53.4 & 54.0 & 55.0 & 55.5 \\
Min-K\%++
& 49.3 & 50.9 & \uline{53.0} & \uline{53.6} & \textbf{56.2}
& \textbf{50.1} & \textbf{50.2} & \textbf{50.2} & \textbf{50.5} & \textbf{50.4}
& \uline{50.7} & \uline{51.3} & \textbf{52.6} & \uline{54.1} & \textbf{55.8}
& 52.0 & \uline{53.7} & \uline{54.5} & \uline{55.9} & \uline{57.1} \\
\rowcolor{gray!15}
Gap-K\%
& 49.9 & \textbf{51.5} & \textbf{53.3} & \textbf{53.8} & \textbf{56.2}
& \uline{50.0} & \uline{49.9} & \uline{49.9} & \uline{50.2} & \uline{50.1}
& 50.2 & 51.2 & 52.2 & \textbf{54.2} & \uline{55.6}
& 52.1 & \textbf{53.9} & \textbf{54.6} & \textbf{56.1} & \textbf{57.3} \\
\bottomrule
\end{tabular}
\end{table*}

%% file: Figures/Figure3.tex
\begin{figure}[t]
  \centering
  \includegraphics[width=\linewidth]{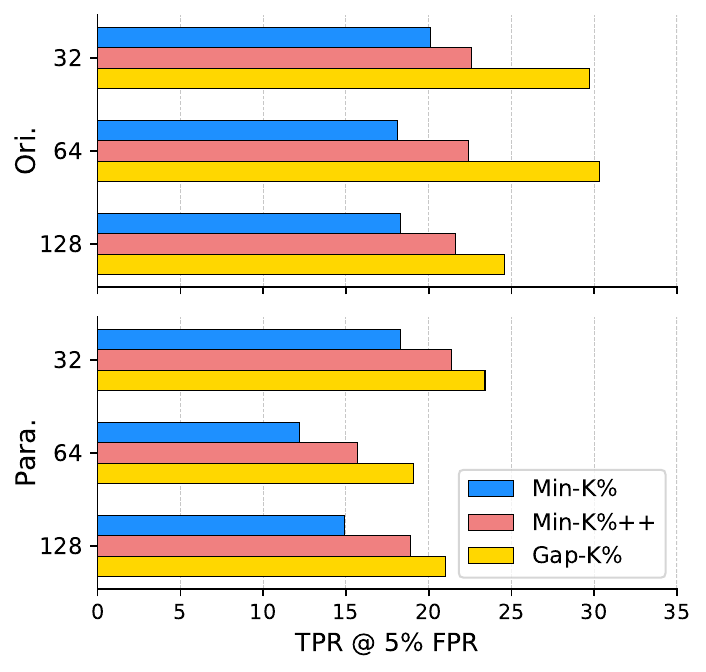}
  \vspace{-0.1in}
  \caption{Average TPR@5\%FPR across five models on WikiMIA, comparing Min-K\%, Min-K\%++, and Gap-K\%. Results are shown for original and paraphrased inputs at sequence lengths of 32, 64, and 128.
  Full results are reported in Appendix \ref{app:fpr}.}
  \vspace{-0.15in}
  \label{fig:fig3}
\end{figure}

%% file: Tables/Table3.tex
\begin{table}[t]
\centering
\caption{Fraction of tokens whose normalized gap magnitude exceeds threshold $\tau$.}
\begin{tabular}{c|cc}
\toprule
Threshold $\tau$ & Train & Non-train \\
\midrule
1 & 0.7244 & 0.7400 \\
2 & 0.5063 & 0.5414 \\
3 & 0.3553 & 0.3994 \\
4 & 0.2478 & 0.2924 \\
5 & 0.1638 & 0.2037 \\
6 & 0.0998 & 0.1333 \\
\bottomrule
\end{tabular}
\label{tab:gap_threshold}
\end{table}

%% file: Figures/Figure4.tex
\begin{figure}[t]
  \centering
  \includegraphics[width=\linewidth]{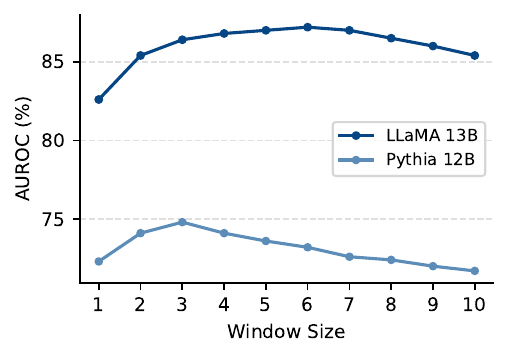}
  \vspace{-0.2in}
  \caption{Effect of window size on sequential smoothing for LLaMA-13B and Pythia-12B.}
  \label{fig:window_size}
\end{figure}

%% file: Tables/Table4.tex
\begin{table}[t]
\centering
\caption{Effect of sequential locality. Applying sequential smoothing improves performance only when the original token order is preserved.}
\setlength{\tabcolsep}{3.5pt}
\label{tab:sequential_locality}
\begin{tabular}{l|c}
\toprule
Method & AUROC (\%) \\
\midrule
No smoothing & 72.3 \\
Shuffled-order smoothing & 72.9 \\
Sequential smoothing (Ours) & 74.8 \\
\bottomrule
\end{tabular}
\end{table}

%% file: Tables/Table5.tex
\begin{table}[t]
\centering
\small
\caption{Ablation of key components relative to Min-K\%++. Top-1 denotes replacing the mean in Min-K\%++ (Eq.~\ref{eq:minkpp_token}) with the top-1 (maximum) log probability, and smoothing denotes applying sequential smoothing.}
\label{tab:ablation}
\begin{tabular}{@{}l|cc|c@{}}
\toprule
Method & Top-1 & Smoothing & AUROC (\%) \\
\midrule
Min-K\%++        &        &        & 72.6 \\
\quad + Top-1    & \checkmark &        & 72.3 \\
\quad + Smoothing   &        & \checkmark & 73.8 \\
Gap-K\% (Ours)            & \checkmark & \checkmark & 74.8 \\
\bottomrule
\end{tabular}
\end{table}

%% file: Figures/Figure5.tex
\begin{figure}[t]
  \centering
  \small
  \includegraphics[width=\linewidth]{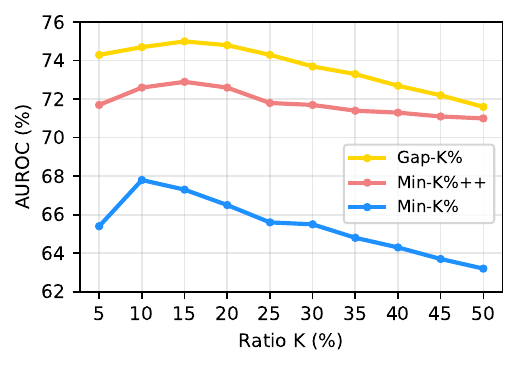}
  \vspace{-0.2in}
  \caption{Effect of the $k\%$ ratio on AUROC.}
  \label{fig:ratio}
  \vspace{-0.1in}
\end{figure}

%% file: Figures/Figure6.tex
\begin{figure*}[t]
  \centering
  \includegraphics[width=\linewidth]{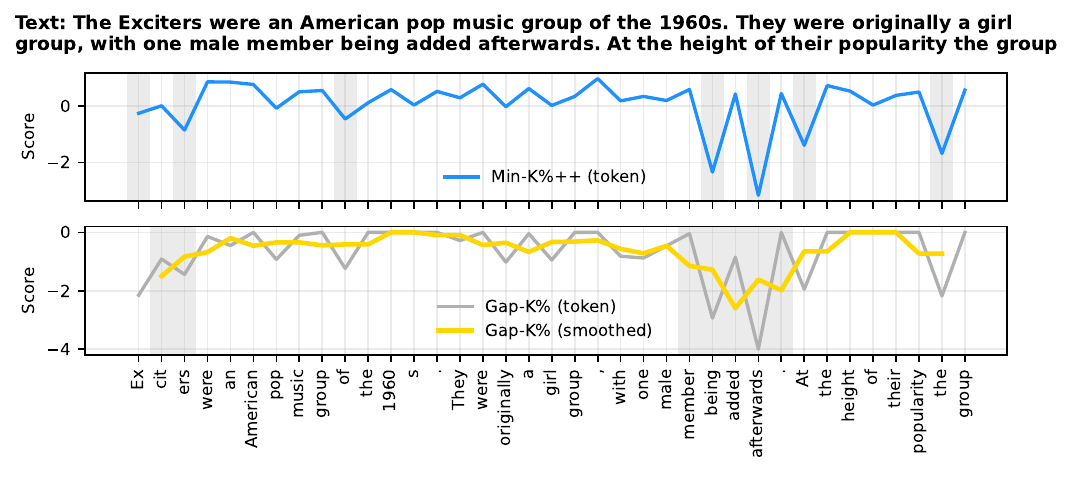}
  \vspace{-0.2in}
  \caption{Visualization of token-level scores for Min-K\%++ and Gap-K\% on a representative WikiMIA-32 sample using Pythia-12B.
Min-K\%++ (top) selects tokens based on token-level scores, with shaded regions denoting the bottom 20\% of tokens. Gap-K\% (bottom) applies sequential smoothing to token-level scores and selects the bottom 20\% based on the resulting window-averaged scores, which are indicated by shaded regions.}
  \label{fig:token_analysis}
  \vspace{-0.1in}
\end{figure*}

%% file: Tables/Table6.tex
\begin{table}[t]
\centering
\small
\caption{AUROC results on recent state-of-the-art LLMs (LLaMA 3.1 and Gemma2) and their instruction-tuned variants, evaluated on WikiMIA-25.}
\setlength{\tabcolsep}{1.5pt}
\begin{tabular}{@{}lcccc@{}}
\toprule
Method 
& \begin{tabular}{c}LLaMA 3.1\\8B\end{tabular}
& \begin{tabular}{c}LLaMA 3.1\\8B Instr.\end{tabular}
& \begin{tabular}{c}Gemma2\\9B\end{tabular}
& \begin{tabular}{c}Gemma2\\9B Instr.\end{tabular} \\
\midrule
Loss        & 66.9 & 64.9 & 63.3 & 61.4 \\
Zlib        & 67.3 & 65.3 & 63.7 & 61.6 \\
Neighbor      & 69.2 & 67.0 & 65.6 & 61.8 \\
Min-K\%     & 71.1 & 68.4 & 65.1 & 64.3 \\
Min-K\%++     & 82.7 & 73.1 & 75.6 & 64.5 \\
Gap-K\% & \textbf{84.1} & \textbf{76.6} & \textbf{78.3} & \textbf{65.8} \\
\bottomrule
\end{tabular}
\label{tab:recent_models}
\end{table}

%% file: Tables/Table7.tex
\begin{table}[t]
\centering
\small
\caption{AUROC results under adversarial paraphrasing using DIPPER.}
\setlength{\tabcolsep}{17pt}
\begin{tabular}{l|c}
\toprule
Method & AUROC (\%) \\
\midrule
Loss        & 52.3 \\
Zlib        & 50.0 \\
Neighbor    & 60.3 \\
Min-K\%     & 57.9 \\
Min-K\%++     & 65.5 \\
Gap-K\% & \textbf{66.6} \\
\bottomrule
\end{tabular}
\label{tab:dipper}
\end{table}

%% file: 5_conclusion.tex
\section{Conclusion} 
In this work, we studied pretraining data detection through the lens of the optimization dynamics of autoregressive large language models (LLMs). 
By analyzing the gradient behavior induced by the next-token prediction objective, we identified discrepancies between top-1 predictions and observed tokens as a meaningful signal for membership inference. 
Based on this insight, we proposed Gap-K\%, a simple yet effective detection method that exploits top-1 log probability gaps and incorporates sequential smoothing to capture local correlations.
Extensive experiments on the WikiMIA and MIMIR benchmarks demonstrate that Gap-K\% consistently outperforms prior state-of-the-art methods. 
Overall, our findings highlight the importance of top-1 prediction behavior, together with sequential smoothing to capture local correlations. 
We hope this perspective motivates further exploration for understanding memorization, privacy risks in LLMs.

\section*{Limitations}
Our method operates under a gray-box assumption, requiring access to token-level probability outputs from the target model. 
While this setting is common in prior work on pretraining data detection and is also assumed by all baselines in our evaluation, it restricts the applicability of our approach to models or APIs that explicitly expose such information. Extending such approaches to fully black-box settings is an important direction for future work.

Although we evaluate Gap-K\% on recent model families including LLaMA 3.1 and Gemma2 with their instruction-tuned variants, a number of other recently released LLM families are not covered in our experiments. In addition, our evaluation is limited to models up to tens of billions of parameters, leaving the behavior of Gap-K\% at the hundreds-of-billions scale unverified. A broader investigation across diverse model families and scales would further clarify the generality of our findings.

Finally, while we assess robustness under strong paraphrasing transformations using DIPPER, these evaluations do not fully capture adaptive adversarial scenarios in which paraphrases are explicitly optimized to evade pretraining data detection. In such settings, an adversary may exploit knowledge of the detection mechanism to manipulate token-level statistics, and evaluating robustness against such detection-aware attacks remains an open challenge.

\section*{Ethical Statements}
This work is conducted with careful attention to ethical research practices and responsible use of data and models. 
The primary objective of our study is to improve the understanding of model behavior and privacy-related risks in LLMs.
All experiments in this study are performed using publicly available datasets and pretrained models that are released for research purposes under permissive open-source licenses. 
Based on the available documentation provided by the dataset and model developers, the resources used in this work do not contain personally identifiable information (PII) and are commonly adopted benchmarks within the research community.

While our methods aim to enhance transparency and facilitate the detection of potential privacy risks in pretrained models, we acknowledge that techniques could be misused if deployed irresponsibly. 
In particular, inferring characteristics of training data may pose risks when applied to sensitive or unauthorized data sources. 
To mitigate such concerns, we emphasize that our approach is intended strictly for non-commercial, academic research. 
Overall, this work aims to contribute to ongoing discussions on transparency and privacy considerations in LLMs.

\section*{Acknowledgments}

All authors are affiliated with the Department of Artificial Intelligence at Yonsei University.
This research was supported in part by Institute for Information \& communications Technology Planning \& Evaluation (IITP) grant funded by the Korea government (MSIT) (No. RS-2020-II201361, Artificial Intelligence Graduate School Program (Yonsei University); No. RS-2025-02215344, Development of AI Technology with Robust and Flexible Resilience Against Risk Factors.

%% file: 6_appendix.tex
\input{Tables/Table8}
\input{Tables/Table9}

\newpage

\section{Full TPR@5\%FPR Results}
\label{app:fpr}
We report full TPR@5\%FPR results on the WikiMIA and MIMIR benchmarks in Tables~\ref{tab:fpr_wikimia} and \ref{tab:fpr_mimir}, respectively. 
TPR@5\%FPR evaluates performance at the extreme tail of the score distribution, where only a small number of samples determine the outcome. 
As a result, this metric is inherently sensitive to variance across datasets and models, which can make consistent trends less apparent.
Similar fluctuations have been observed in prior work ~\citep{zhang2025mink}, suggesting that this behavior is inherent to the metric.
For WikiMIA, Gap-K\% consistently outperforms Min-K\%++ across different input lengths.
In the original setting, averaged over five models, Gap-K\% improves upon Min-K\%++ by 7.1\%, 7.9\%, and 3.0\% for input lengths of 32, 64, and 128, respectively.
In the paraphrased setting, Gap-K\% also achieves gains of 2.0\%, 3.4\%, and 2.1\% over Min-K\%++ for the same input lengths.
These results further demonstrate the robustness of Gap-K\% across both original and paraphrased settings. 
For MIMIR, no single method consistently achieves the best performance across all model sizes. 
Despite this, Gap-K\% attains the highest average performance in the Pythia-6.9B setting and achieves the second-best average scores for the 1.4B, 2.8B, and 12B models.

\input{Tables/Table10}
\input{Tables/Table11}
\input{Figures/Figure7}
\section{Understanding Difficulty of Pile-CC}
\label{app:pile-cc}
We further investigate why detection methods struggle on the Pile-CC subset of MIMIR.
To better understand this behavior, we analyze the entropy of the next-token distribution using Pythia-2.8B.
As shown in Table~\ref{tab:entropy}, Pile-CC exhibits the highest average entropy among all domains.
This suggests that the domain is highly heterogeneous and distributionally broad, with substantial artifacts and formatting noise in pretraining web corpora. 
As a result, the model struggles to assign high probability to the target token even for training samples, leading to weaker memorization signals and limiting the separability for all detection methods.

\section{Effect of Window Size across Model Families}
\label{app:window_size}
We investigate why the optimal window size differs across model families.
We hypothesize that this difference may arise from either training procedures or architectural characteristics.
To disentangle these factors, we evaluate models that share architectural similarities with LLaMA but differ in training methodology, including Mistral-7B v0.1 \citep{jiang2023mistral7b} and OpenLLaMA-3B \citep{openlm2023openllama}.
Since the original WikiMIA is not suitable for these newer models due to potential temporal overlap with their training data, we instead evaluate on WikiMIA-25 (length 64 split).
As shown in Table~\ref{tab:window_size}, all models consistently achieve their best performance at window size 6.
We further confirm that LLaMA-13B exhibits the same optimal window size under WikiMIA-25.
These results suggest that the observed differences in optimal window size are more likely attributable to architectural factors rather than training procedures.

\input{Figures/Figure8}
\section{Window Size for Other Models}
We further evaluate the effect of window size across additional models.
Following the experimental setup in Section \ref{sec:4.3}, we use the WikiMIA-64 dataset. 
Due to the high computational cost, we do not include results for LLaMA-65B.
Fig.~\ref{fig:app_fig7} shows that both Mamba-1.4B and Pythia-6.9B achieve their best performance at a window size of 3, consistent with the results for Pythia-12B.

\input{Tables/Table12}

\section{Effect of Token Selection Strategy}
\label{app:token_selection}
We analyze different token selection strategies to validate our choice of selecting low-scoring tokens.
As shown in Table~\ref{tab:token_selection}, selecting the bottom $k$\% tokens (Bottom-K\%) achieves the highest AUROC.
In contrast, selecting high-scoring tokens (Top-K\%) or random tokens leads to substantially worse results.
Using all tokens without $k$\% selection also underperforms compared to Bottom-K\%, highlighting the importance of focusing on informative tokens.
Notably, low-scoring tokens tend to correspond to those with larger gaps between the target and top-1 predictions, making them particularly informative under our gap-based criterion.

\section{Histogram-Based Comparison of Training and Non-Training Samples}
Fig.~\ref{fig:app_fig8} visualizes the distributions of the proposed Gap-K\% scores for member and non-member samples. 
We plot the results for Pythia-12B and LLaMA-13B on the WikiMIA-64 benchmark. 
The histograms illustrate the degree of separation induced by the score.

\section{Additional Token-level Visualization}
\input{Figures/Figure9}
Fig.~\ref{fig:app_fig9} presents additional qualitative examples of token-level score visualizations for Min-K\%++ and Gap-K\%.
Consistent with the observation in Fig.~\ref{fig:token_analysis} in the main text, Min-K\%++ selects isolated low-scoring tokens across the sequence, whereas Gap-K\% identifies contiguous low-score regions after applying sequential smoothing.

\section{Usage of AI assistants}
During the preparation of this paper, AI-assisted writing tools were used for editorial purposes, including improving readability, coherence, and grammatical accuracy. 
Their use was restricted to language polishing, without generating or altering any technical content. 
All aspects of the method and experimental outcomes were developed independently by the authors. 
The application of AI assistance was limited to surface-level revisions and did not affect the originality or scientific substance of the work.

%% file: Tables/Table8.tex
\begin{table*}[t]
\centering
\scriptsize
\setlength{\tabcolsep}{8pt}
\renewcommand{\arraystretch}{1.15}

\caption{TPR@5\%FPR results on WikiMIA. The best and second-best scores are highlighted in \textbf{bold} and \uline{underlined}, respectively.}
\label{tab:fpr_wikimia}

\begin{tabular}{c l cc cc cc cc cc cc}
\toprule
& & \multicolumn{2}{c}{\textbf{Mamba-1.4B}}
  & \multicolumn{2}{c}{\textbf{Pythia-6.9B}}
  & \multicolumn{2}{c}{\textbf{Pythia-12B}}
  & \multicolumn{2}{c}{\textbf{LLaMA-13B}}
  & \multicolumn{2}{c}{\textbf{LLaMA-65B}}
  & \multicolumn{2}{c}{\textbf{Average}} \\
\cmidrule(lr){3-4}\cmidrule(lr){5-6}\cmidrule(lr){7-8}\cmidrule(lr){9-10}\cmidrule(lr){11-12}\cmidrule(lr){13-14}
\textbf{Len.} & \textbf{Method}
& \textit{Ori.} & \textit{Para.}
& \textit{Ori.} & \textit{Para.}
& \textit{Ori.} & \textit{Para.}
& \textit{Ori.} & \textit{Para.}
& \textit{Ori.} & \textit{Para.}
& \textit{Ori.} & \textit{Para.} \\
\midrule

% ===================== Len 32 =====================
\multirow{6}{*}{32}
& Loss
& 14.2 & \textbf{14.2}
& 14.2 & 15.0
& 17.1 & 17.3
& 14.0 & \uline{16.3}
& 22.0 & 21.7
& 16.3 & 16.9 \\
& Zlib
& \uline{15.5} & 13.2
& 16.3 & 12.7
& 17.1 & 15.5
& 11.6 & 15.0
& 19.6 & 17.3
& 16.0 & 14.7 \\
& Neighbor
& 11.9 & 7.2
& \uline{16.5} & 9.6
& 19.4 & 9.8
& 11.6 & 8.5
& 6.5 & 12.1
& 13.2 & 9.4 \\
& Min-K\%
& 14.2 & 11.9
& \textbf{17.8} & \textbf{21.7}
& \uline{23.0} & \textbf{19.9}
& 18.9 & 14.2
& 26.4 & 23.8
& 20.1 & 18.3 \\
& Min-K\%++
& 11.4 & 7.8
& 14.5 & 14.5
& 16.5 & 15.5
& \uline{33.1} & \textbf{33.9}
& \uline{37.5} & \textbf{35.4}
& \uline{22.6} & \uline{21.4} \\
\rowcolor{gray!15}
& Gap-K\%
& \textbf{17.3} & \uline{13.4}
& 16.3 & \uline{18.9}
& \textbf{24.0} & \uline{19.4}
& \textbf{43.4} & \textbf{33.9}
& \textbf{47.3} & \uline{31.3}
& \textbf{29.7} & \textbf{23.4} \\
\midrule

% ===================== Len 64 =====================
\multirow{6}{*}{64}
& Loss
& 9.5 & 8.1
& 13.4 & 10.6
& 9.2 & 11.6
& 11.3 & 12.0
& 15.5 & 13.4
& 11.8 & 11.1 \\
& Zlib
& 14.1 & \textbf{15.1}
& 16.2 & \textbf{15.8}
& 11.3 & \textbf{16.2}
& 12.7 & 13.4
& 17.6 & 18.3
& 14.4 & \uline{15.8} \\
& Neighbor
& 8.8 & 9.5
& 10.9 & 12.7
& 11.3 & 10.6
& 10.2 & 14.4
& 9.9 & 16.9
& 10.2 & 12.8 \\
& Min-K\%
& \uline{15.8} & 7.7
& \uline{19.0} & 12.7
& \uline{21.5} & 12.7
& 17.3 & 13.4
& 16.9 & 14.4
& 18.1 & 12.2 \\
& Min-K\%++
& 13.7 & 7.0
& 16.2 & 10.2
& 16.9 & 10.9
& \uline{31.3} & \uline{23.2}
& \uline{33.8} & \uline{27.1}
& \uline{22.4} & 15.7 \\
\rowcolor{gray!15}
& Gap-K\%
& \textbf{20.1} & \uline{10.9}
& \textbf{23.6} & \uline{13.0}
& \textbf{25.4} & \uline{13.7}
& \textbf{44.7} & \textbf{30.3}
& \textbf{37.7} & \textbf{27.8}
& \textbf{30.3} & \textbf{19.1} \\
\midrule

% ===================== Len 128 =====================
\multirow{6}{*}{128}
& Loss
& 11.5 & \uline{13.7}
& 14.4 & 16.5
& 18.0 & \textbf{19.4}
& 21.6 & 18.0
& 20.1 & 24.5
& 17.1 & 18.4 \\
& Zlib
& \textbf{19.4} & \textbf{17.3}
& \textbf{20.9} & \textbf{20.9}
& \textbf{23.7} & \textbf{19.4}
& 18.7 & 21.6
& 23.0 & 22.3
& 21.1 & \uline{20.3} \\
& Neighbor
& 15.8 & \uline{13.7}
& 10.8 & 17.3
& 10.1 & 10.1
& 12.9 & 13.7
& 15.8 & 18.7
& 13.1 & 14.7 \\
& Min-K\%
& 9.4 & 5.0
& 18.0 & 16.5
& 20.1 & \uline{18.7}
& 20.1 & 15.1
& \uline{23.7} & 19.4
& 18.3 & 14.9 \\
& Min-K\%++
& 10.1 & 6.5
& \uline{20.1} & \uline{18.0}
& 18.0 & 9.4
& \uline{38.1} & \textbf{35.3}
& 21.6 & \uline{25.2}
& \uline{21.6} & 18.9 \\
\rowcolor{gray!15}
& Gap-K\%
& \uline{16.5} & \uline{13.7}
& 19.4 & 13.7
& \uline{21.6} & 18.0
& \textbf{39.6} & \uline{33.1}
& \textbf{25.9} & \textbf{26.6}
& \textbf{24.6} & \textbf{21.0} \\
\bottomrule
\end{tabular}
\end{table*}

%% file: Tables/Table9.tex
\begin{table*}[t!]
\centering
\scriptsize
\setlength{\tabcolsep}{3.2pt}
\renewcommand{\arraystretch}{1.15}

\caption{TPR@5\%FPR results on MIMIR. The best and second-best scores are highlighted in \textbf{bold} and \uline{underlined}, respectively.}
\label{tab:fpr_mimir}

\begin{tabular}{l ccccc ccccc ccccc ccccc}
\toprule
& \multicolumn{5}{c}{\textbf{Wikipedia}}
& \multicolumn{5}{c}{\textbf{Github}}
& \multicolumn{5}{c}{\textbf{Pile CC}}
& \multicolumn{5}{c}{\textbf{PubMed Central}} \\
\cmidrule(lr){2-6}\cmidrule(lr){7-11}\cmidrule(lr){12-16}\cmidrule(lr){17-21}
\textbf{Method}
& 160M & 1.4B & 2.8B & 6.9B & 12B
& 160M & 1.4B & 2.8B & 6.9B & 12B
& 160M & 1.4B & 2.8B & 6.9B & 12B
& 160M & 1.4B & 2.8B & 6.9B & 12B \\
\midrule
Loss
& 4.2 & 4.7 & 4.7 & 5.1 & 5.0
& 22.6 & 32.1 & 33.6 & 38.5 & 30.3
& 3.1 & \uline{5.0} & \uline{4.8} & 4.9 & 5.1
& 4.0 & 4.4 & 4.3 & 4.9 & 5.0 \\
Zlib
& 4.2 & \textbf{5.7} & \uline{5.9} & 6.3 & 6.8
& \uline{25.1} & \uline{32.8} & \textbf{36.2} & \textbf{40.1} & 32.9
& \uline{4.0} & \textbf{5.1} & \textbf{5.4} & \textbf{6.2} & \textbf{6.6}
& 3.8 & 3.6 & 3.5 & 4.3 & 4.4 \\
Neighbor
& 4.0 & 4.5 & 4.9 & 5.8 & /
& 24.7 & 31.6 & 29.8 & 34.1 & /
& 3.9 & 3.6 & 4.0 & 5.3 & /
& 3.9 & 3.7 & 4.5 & 4.5 & / \\
Min-K\%
& 4.8 & \uline{5.6} & 5.0 & 6.1 & 5.8
& 22.6 & 31.5 & 34.0 & \uline{39.0} & 32.6
& 3.5 & 4.5 & \uline{4.8} & 5.0 & 4.8
& 4.7 & 4.6 & 4.5 & 5.1 & 4.9 \\
Min-K\%++
& \uline{5.2} & 5.3 & \uline{5.9} & \uline{7.0} & \textbf{7.8}
& \textbf{25.2} & \textbf{33.0} & 34.2 & 38.2 & \uline{34.5}
& \textbf{5.0} & 3.7 & 3.7 & 4.8 & 4.6
& \uline{4.8} & \textbf{6.1} & \uline{4.8} & \uline{5.6} & \textbf{6.4} \\
\rowcolor{gray!15}
Gap-K\%
& \textbf{5.5} & 5.4 & \textbf{6.0} & \textbf{7.5} & \uline{7.0}
& 22.9 & 32.3 & \uline{34.3} & 38.5 & \textbf{34.8}
& 3.7 & 4.6 & 4.3 & \uline{5.8} & \uline{5.9}
& \textbf{5.4} & \uline{5.2} & \textbf{5.4} & \textbf{5.7} & \uline{5.8} \\
\midrule

& \multicolumn{5}{c}{\textbf{ArXiv}}
& \multicolumn{5}{c}{\textbf{DM Mathematics}}
& \multicolumn{5}{c}{\textbf{HackerNews}}
& \multicolumn{5}{c}{\textbf{Average}} \\
\cmidrule(lr){2-6}\cmidrule(lr){7-11}\cmidrule(lr){12-16}\cmidrule(lr){17-21}
\textbf{Method}
& 160M & 1.4B & 2.8B & 6.9B & 12B
& 160M & 1.4B & 2.8B & 6.9B & 12B
& 160M & 1.4B & 2.8B & 6.9B & 12B
& 160M & 1.4B & 2.8B & 6.9B & 12B \\
\midrule
Loss
& 4.0 & \textbf{4.8} & 4.6 & 5.4 & 5.6
& 3.8 & 4.3 & 4.1 & 4.1 & 4.0
& \uline{5.0} & 4.8 & 5.5 & 5.9 & \textbf{6.8}
& 6.7 & 8.6 & 8.8 & 9.8 & 8.8 \\
Zlib
& 2.9 & 4.3 & 4.1 & 4.6 & 4.7
& 4.1 & \uline{5.0} & 4.6 & 4.3 & 4.3
& \uline{5.0} & \textbf{5.5} & \textbf{5.8} & 5.6 & 5.8
& 7.0 & \textbf{8.9} & \textbf{9.4} & 10.2 & 9.4 \\
Neighbor
& 4.7 & \textbf{4.8} & 4.4 & 4.1 & /
& \textbf{5.6} & 4.4 & 4.5 & \uline{4.5} & /
& \textbf{6.5} & \uline{5.2} & 5.3 & 5.7 & /
& \uline{7.6} & 8.3 & 8.2 & 9.1 & / \\
Min-K\%
& 4.4 & 4.3 & 4.5 & 5.4 & 5.3
& 3.9 & 4.1 & 4.6 & 4.3 & \uline{4.6}
& 4.2 & 4.6 & \uline{5.7} & \textbf{6.3} & \uline{6.1}
& 6.9 & 8.5 & 9.0 & 10.2 & 9.2 \\
Min-K\%++
& \textbf{5.4} & \uline{4.7} & \textbf{6.1} & \textbf{6.8} & \textbf{7.0}
& \uline{4.4} & 4.8 & \textbf{5.4} & \uline{4.5} & \textbf{5.4}
& 4.4 & 3.5 & 4.6 & 5.7 & 5.7
& \textbf{7.8} & \uline{8.7} & 9.2 & \uline{10.4} & \textbf{10.2} \\
\rowcolor{gray!15}
Gap-K\%
& \uline{5.2} & 4.1 & \uline{5.3} & \uline{6.4} & \uline{6.1}
& 4.2 & \textbf{5.2} & \uline{4.8} & \textbf{5.3} & \textbf{5.4}
& 3.8 & 3.9 & 5.1 & \uline{6.2} & 5.1
& 7.2 & \uline{8.7} & \uline{9.3} & \textbf{10.8} & \uline{10.0} \\
\bottomrule
\end{tabular}
\end{table*}

%% file: Tables/Table10.tex
\begin{table*}[t!]
\centering
\caption{Average entropy of next-token distributions across MIMIR subsets for training and non-training samples. Pile-CC exhibits the highest entropy among all domains.}
\setlength{\tabcolsep}{3pt}
\begin{tabular}{lccccccc}
\toprule
 & ArXiv & DM Mathematics & Github & HackerNews & Pile CC & PubMed Central & Wikipedia \\
\midrule
Train     & 2.05 & 1.28 & 0.82 & 2.51 & 2.59 & 2.00 & 2.01 \\
Not Train & 2.05 & 1.27 & 1.22 & 2.51 & 2.58 & 1.99 & 2.01 \\
\bottomrule
\end{tabular}
\label{tab:entropy}
\end{table*}

%% file: Tables/Table11.tex
\begin{table*}[t!]
\centering
\caption{AUROC (\%) across different window sizes for models with LLaMA-based architectures. All models consistently achieve their best performance at window size 6.}
\setlength{\tabcolsep}{13pt}
\begin{tabular}{lccccccc}
\toprule
\multirow{2}{*}{Model} & \multicolumn{7}{c}{Window Size} \\
\cmidrule(lr){2-8}
      & 2 & 3 & 4 & 5 & 6 & 7 & 8 \\
\midrule
OpenLLaMA-3B & 77.4 & 78.4 & 78.7 & \textbf{78.9} & \textbf{78.9} & 78.6 & 77.8 \\
Mistral-7B   & 86.3 & 87.8 & 88.7 & 89.3 & \textbf{90.0} & 89.3 & 88.4 \\
LLaMA-13B    & 81.3 & 82.2 & 83.0 & 83.4 & \textbf{83.9} & 83.8 & 83.5 \\
\bottomrule
\end{tabular}
\label{tab:window_size}
\end{table*}

%% file: Figures/Figure7.tex
\begin{figure*}[t!]
  \centering
  \includegraphics[width=\linewidth]{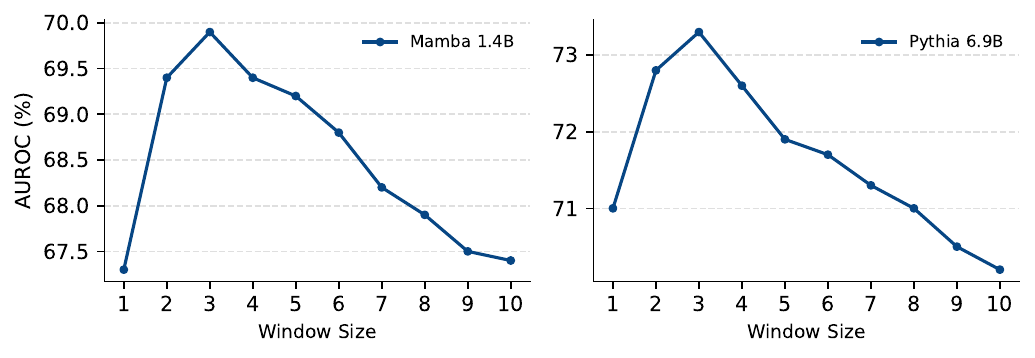}
  \caption{Effect of window size on sequential smoothing
for Mamba-1.4B and Pythia-6.9B.}
  \label{fig:app_fig7}
  \vspace{-0.1in}
\end{figure*}

%% file: Figures/Figure8.tex
\begin{figure*}[t]
  \centering
  \includegraphics[width=0.49\linewidth]{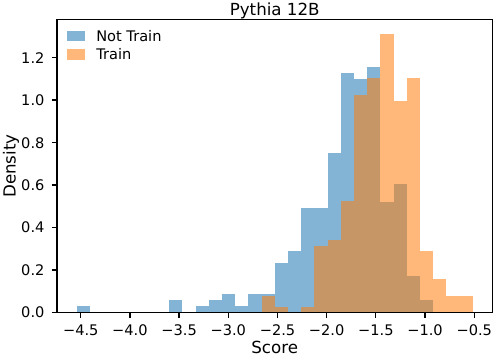}\hfill
  \includegraphics[width=0.49\linewidth]{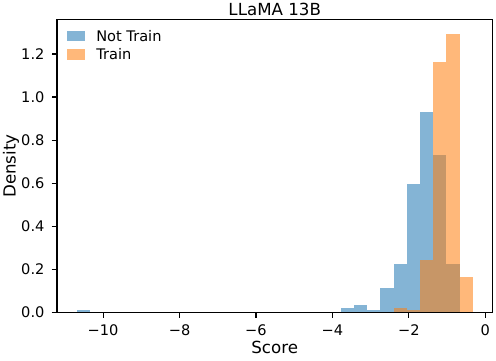}
  \caption{Histogram of the Gap-K\% score distributions for trained and not trained samples on the WikiMIA-64 benchmark. The left shows results from Pythia-12B, and the right shows results from LLaMA-13B.}
  \label{fig:app_fig8}
\end{figure*}

%% file: Tables/Table12.tex
\begin{table}[t]
\centering
\caption{Effect of token selection strategy.}
\setlength{\tabcolsep}{5pt}
\begin{tabular}{l|c}
\toprule
Token Selection & AUROC (\%) \\
\midrule
All tokens (no $k$\% selection) & 68.6 \\
Random-K\%        & 62.6 \\
Top-K\%           & 54.2 \\
Bottom-K\% (Ours) & \textbf{74.8} \\
\bottomrule
\end{tabular}
\label{tab:token_selection}
\end{table}

%% file: Figures/Figure9.tex
\begin{figure*}[t]
  \centering
  \includegraphics[width=\linewidth]{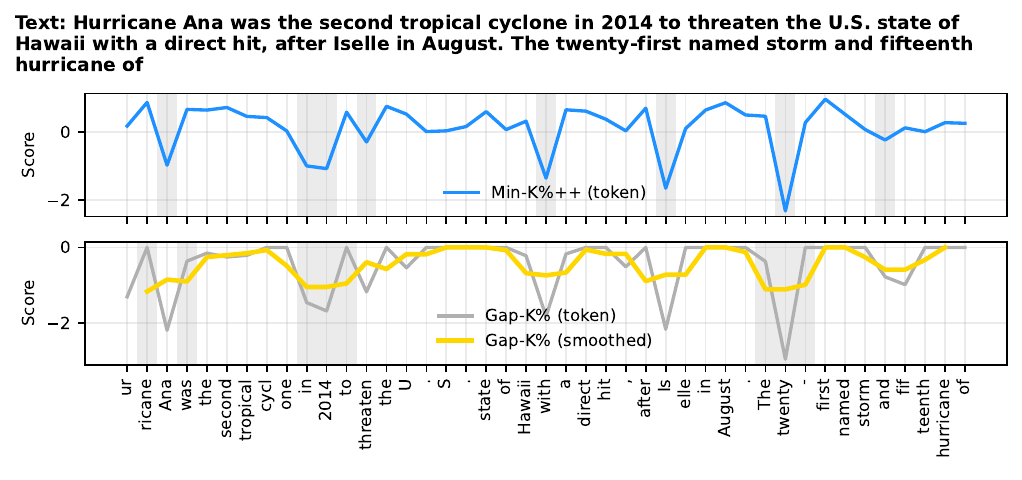}
  \includegraphics[width=\linewidth]{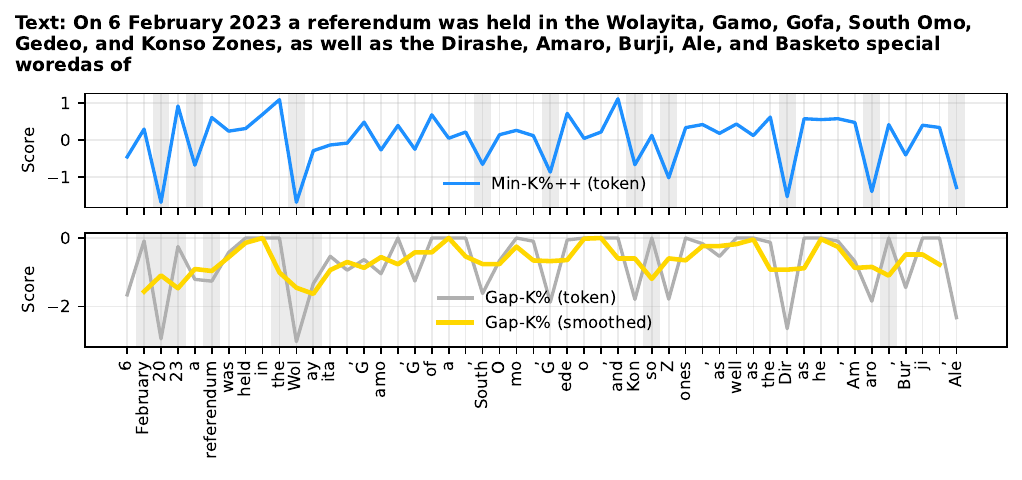}
  \includegraphics[width=\linewidth]{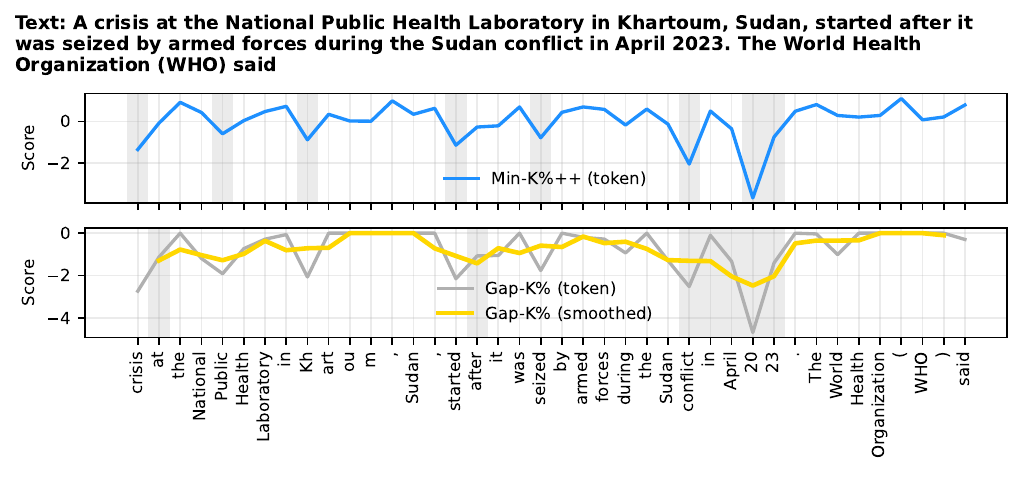}
  \caption{Additional examples of token-level score visualizations for Gap-K\% and Min-K\%++.}
  \label{fig:app_fig9}
\end{figure*}